\documentclass[12pt,letterpaper,twoside]{article}

\usepackage[dvips]{graphicx}
\usepackage{url,times,wrapfig}
\usepackage[utf8]{inputenc}
\usepackage[english, frenchb]{babel}

\usepackage{amsmath}
\usepackage{amssymb}
\usepackage{amscd}
\usepackage{url}
\usepackage{caption}
\usepackage{subfig}

\usepackage{array,supertabular}
\newcolumntype{M}[1]{>{\centering}m{#1}}

\usepackage{taln2010}

\newcommand{\saturated}{{\ensuremath{\leftrightarrow}}}
\newcommand{\positive}{{\ensuremath{\rightarrow}}}
\newcommand{\negative}{{\ensuremath{\leftarrow}}}
\newcommand{\virtual}{{\ensuremath{\sim}}}

\newcommand{\D}{{\mathcal{D}}} 
\newcommand{\T}{{\mathcal{T}}} 
\newcommand{\G}{{\mathcal{G}}} 
\newcommand{\I}{{\mathcal{I}}}

\newcommand{\french}[1]{{\emph{``#1''}}}

\newcommand{\node}[1]{\ensuremath{\mathtt {#1}}}
\newcommand{\inode}[1]{\ensuremath{\mathtt {[#1]}}}
\newcommand{\mnode}[1]{\ensuremath{\mathtt {\lbrace#1\rbrace}}}

\newcommand{\grf}{$\G_f$}
\newcommand{\ig}{GI} 

\newcommand{\feat}[1]{{\ensuremath{\mathtt{#1}}}}
\newcommand{\rev}[1]{{\ensuremath{\mathtt{#1}^{-1}}}}

\newcommand{\app}[1]{{\ensuremath{\overline{\node{#1}}}}}

\usepackage{color}
\newcommand{\new}[1]{{#1}}

\title{Motifs de graphe pour le calcul de dépendances syntaxiques complètes}

\author{
  {Jonathan Marchand, Bruno Guillaume,  Guy Perrier\\
  INRIA Nancy-Grand Est - LORIA - Nancy-Université}
}

\begin{document}

\maketitle

\resume{
Cet article propose une méthode pour calculer les dépendances
syntaxiques d'un énoncé à partir du processus d'analyse en
constituants.  L'objectif est d'obtenir des dépendances complètes
c'est-à-dire contenant toutes les informations nécessaires à la construction de
la sémantique.  Pour l'analyse en constituants, on utilise le
formalisme des grammaires d'interaction~: celui-ci place au cœur de
la composition syntaxique un mécanisme de saturation de polarités qui
peut s'interpréter comme la réalisation d'une relation de dépendance.
Formellement, on utilise la notion de motifs de graphes au sens de la
réécriture de graphes pour décrire les conditions nécessaires à la
création d'une dépendance.
}

\abstract{
\selectlanguage{english}
This article describes a method to build syntactical dependencies
starting from the phrase structure parsing process. The goal is to
obtain all the information needed for a detailled semantical
analysis. Interaction Grammars are used for parsing; the saturation of
polarities which is the core of this formalism can be mapped to
dependency relation. Formally, graph patterns are used to express the
set of constraints which control dependency creations.
}

\selectlanguage{french}
\motsClefs{Analyse syntaxique, dépendance, grammaires
d'interaction, polarité}
{Syntactic analysis, dependency, interaction grammars, polarity}

\vspace{-3mm}
\section*{Introduction}
\vspace{-2mm}

Quand on envisage l'analyse syntaxique en vue de produire une analyse
sémantique complète de la phrase, il est intéressant de représenter le
résultat sous forme de dépendances entre mots. On s'abstrait ainsi de
tous les détails qui ne jouent pas de rôle dans le calcul de la sémantique
afin de ne conserver que l'essentiel. Mais alors, il est important de
définir des structures en dépendances suffisamment riches pour
permettre un calcul fin et complet des relations sémantiques. La
campagne {\sc
  Passage}\footnote{\url{http://atoll.inria.fr/passage/index.fr.html}}
d'évaluation des analyseurs syntaxiques du français utilise de façon
essentielle de telles structures en dépendances. Les analyseurs
participants à la campagne  devaient produire à la fois un découpage
des phrases en groupes syntaxiques et une annotation de ces phrases à
l'aide de relations entre groupes ou mots\footnote{De fait, toutes
  les relations
  pouvaient être ramenées à des relations entre mots.}. Une des
difficultés était de produire toutes les relations déterminées par la syntaxe, en
particulier les moins immédiates comme celles concernant les sujets
des infinitifs par exemple. Le guide d'annotation de {\sc
  Passage} n'impose aucune contrainte sur la structure de dépendances obtenue.
De fait, la structure en dépendances obtenue est un graphe qui n'est pas toujours un arbre~; il contient même parfois des cycles.

Il existe deux approches pour obtenir des analyses en dépendances. La
première consiste à les calculer directement. Or, pour des raisons
d'efficacité, les analyseurs qui font cela imposent des contraintes
sur les structures en dépendances produites~\cite{kubler09, Debusmann06}.
Généralement, ils ne produisent que des arbres et ne permettent donc
pas de retrouver toutes les relations nécessaires à la construction
d'une représentation sémantique. La seconde approche consiste à
extraire une analyse en dépendances à partir d'une analyse en
constituants~\cite{rambow97, kuza2002, candito2009}. Les relations de
dépendances sont alors extraites de l'arbre syntagmatique de la
phrase, ce qui n'est pas toujours évident, mais surtout l'information
pour produire certaines relations peut être manquante.

La méthode que nous proposons s'apparente à la seconde approche
dans la mesure où nous utilisons une analyse en constituants.
\new {Cependant, comme ~\shortcite{rambow97} et \shortcite{kahane_candito} l'ont observé dans le cas des TAG
il est souvent utile de ne pas s'appuyer seulement sur le résultat de l'analyse
mais sur le processus d'analyse lui-même, pour produire des
dépendances.}
Notre méthode utilise le cadre des Grammaires
d'interaction~(\ig) et en exploite la spécificité~: l'utilisation de
\emph{polarités} pour guider la composition syntaxique. Nous avions,
dans un précédent article~\cite{marchand09}, montré comment obtenir
une analyse en dépendances en réalisant une dépendance entre deux mots
à chaque fois que des polarités qui étiquetaient les objets lexicaux
correspondants se saturaient. Cette approche nous imposait d'ajouter
une nouvelle polarité au système de polarités des \ig~afin de repérer
les saturations qui ne faisaient que contrôler le contexte des mots
lors de l'analyse et qui provoquaient une sur-génération de relations
de dépendances.

La méthode avait été testée sur une grammaire du français à
relativement large échelle~\cite{perrier07} mais les principes qui
avaient présidé à la construction de cette grammaire n'avaient pas
pris en compte l'objectif d'extraire des dépendances syntaxiques
des analyses, dans la mesure où cet objectif est apparu après que la
grammaire ait été construite.  Récemment, la grammaire a été revue
afin d'intégrer des principes exprimant les dépendances
syntaxiques. Cela a permis de se passer de la nouvelle polarité et a
fait apparaître des régularités structurelles dans la saturation des
polarités donnant lieu à la production de dépendances.  Ces
régularités ont été formalisées à l'aide du concept de \emph{motif de
graphe} (\textit{pattern} en anglais dans l'idée de \textit{pattern
matching}).  Un motif est un ensemble de contraintes qui décrit le
contexte structurel dans lequel deux polarités qui se saturent
réalisent une dépendance syntaxique.  Le processus d'analyse avec les
\ig~ étant formalisé sous forme d'un graphe, les dépendances sont
alors créées par détection de motifs dans ce graphe.

La section~\ref{sec-dep} précise ce qu'on entend par analyse en
dépendances syntaxiques complète. La section~\ref{sec-gi} présente
brièvement le formalisme des \ig~et la section~\ref{sec-prin} décrit
les principes de construction de la grammaire du français qui
permettent d'exprimer les dépendances syntaxiques. Enfin, la
section~\ref{sec-motifs} montre comment les motifs de graphe sont
utilisés pour produire des dépendances.

\vspace{-4mm}
\section{Analyse en dépendances syntaxiques complète}
\label{sec-dep}
\vspace{-2mm}

\begin{figure}[ht]
  \centering
  \subfloat[Partielle]{\includegraphics[scale=.75]{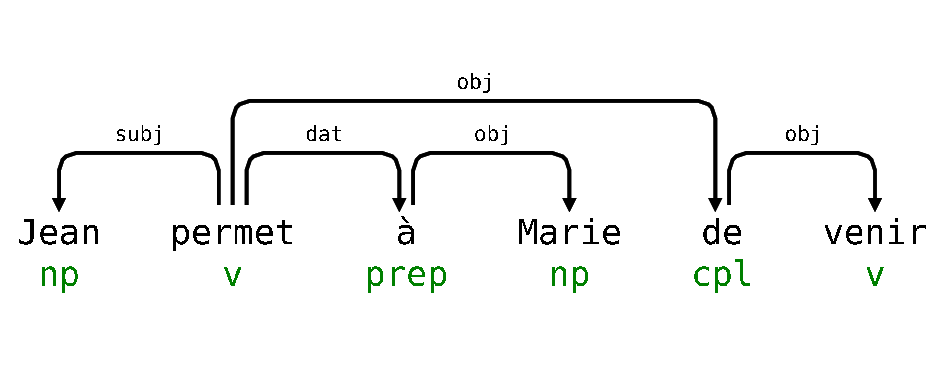}\label{ex-permet-a}}
  \qquad
  \subfloat[Complète]{\includegraphics[scale=.75]{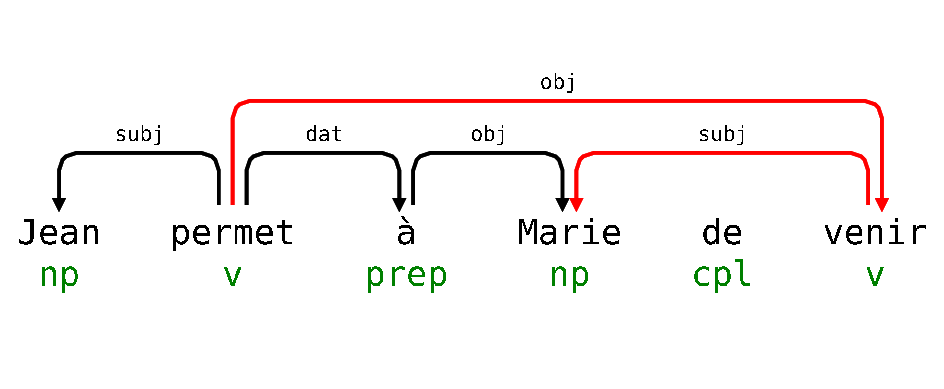}\label{ex-permet-b}}
\caption{\label{ex-permet} Structure en dépendances pour la phrase
  \french{Jean permet à Marie de venir}}
\end{figure}

La notion d'analyse complète fait appel à la différence entre
dépendances dites {\em directes} (en noir dans les figures) et
dépendances {\em indirectes} (en rouge dans les figures), selon qu'elles
se réalisent sans ou à l'aide d'un mot intermédiaire. Dans la
proposition \french{Jean permet à Marie de venir}
(figure~\ref{ex-permet}), \french{Jean} sujet de \french{permet} et
\french{à} complément d'attribution de \french{permet} correspondent à des dépendances
directes~(\ref{ex-permet-a}). La relation \french{Marie} sujet de
\french{venir} est quant à elle une dépendance
indirecte~(\ref{ex-permet-b}).
 
Nous appellerons {\em analyses partielles} les analyses uniquement
composées de dépendances directes. Dans nos exemples, les analyses
partielles sont inspirées par le guide d'annotation de la French
Dependency
Treebank\footnote{\url{http://www.linguist.univ-paris-diderot.fr/~mcandito/Rech/FTBDeps/}}.
Nous appellerons {\em analyses complètes} les analyses qui contiennent
les dépendances indirectes utiles pour l'analyse
sémantique\footnote{Certaines dépendances directes deviennent alors
  inutiles et sont supprimées.}.

Souvent, les dépendances indirectes peuvent être retrouvées à partir
des dépendances directes. Toutefois, ce n'est pas toujours le cas.
Dans la phrase \french{Jean promet à Marie de venir}, la
structure en dépendances partielle est identique à celle de la phrase
\french{Jean permet à Marie de venir}~(\ref{ex-permet-a}). Cependant,
dans la première il y a une dépendance indirecte \french{Jean} sujet
de \french{venir}, et dans la deuxième cette dépendance est entre
\french{Marie} et \french{venir}.

\begin{figure}[ht]
  \centering
  \subfloat[Partielle]{\includegraphics[scale=.75]{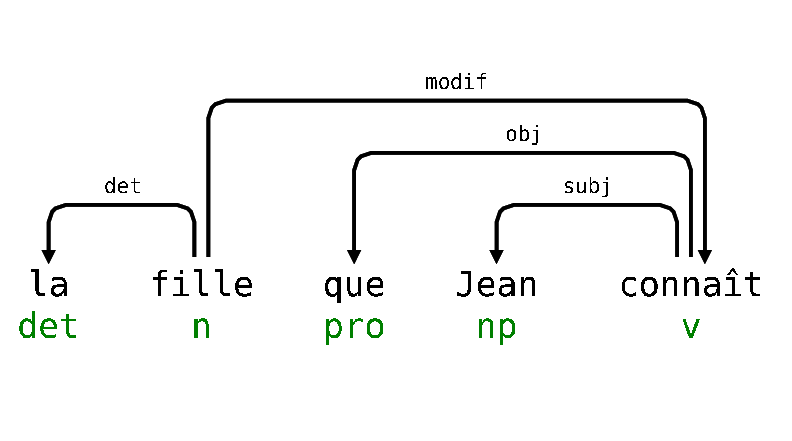}\label{ex-relative-a}}
  \qquad
  \subfloat[Complète]{\includegraphics[scale=.75]{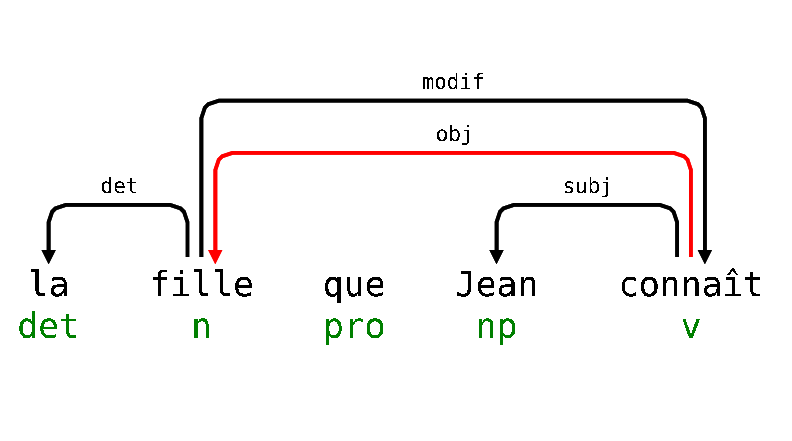}\label{ex-relative-b}}
\caption{\label{ex-relative} Structure en dépendances pour le syntagme
  nominal \french{la fille que Jean connaît}}
\end{figure}

La figure~(\ref{ex-permet-b}) montre déjà que les dépendances 
complètes ne forment pas un arbre. Dans le syntagme
nominal contenant une relative \french{la fille que Jean connaît},
l'analyse complète (figure \ref{ex-relative-b}) n'utilise plus le
pronom relatif \french{que} comme relais pour introduire la relative
et rappeler l'objet de \french{connaît}. La relation \french{fille}
objet de \french{connaît} est une dépendance indirecte qui introduit
un cycle dans la structure. De plus, la structure n'est plus connexe~:
le pronom relatif \french{que} qui a servi d'intermédiaire entre la
relative et son antécédent n'a plus d'utilité.
 
\vspace{-4mm}
\section{Le formalisme des  grammaires d'interaction}
\label{sec-gi}
\vspace{-2mm}

Les grammaires d'interaction~\cite{Per2003} sont un formalisme grammatical qui place la notion de \emph{polarité} au cœur du mécanisme de composition syntaxique.  Les objets de base d'une grammaire d'interaction sont des fragments d'arbres syntaxiques sous-spécifiés qui sont décorés par des polarités. Ces polarités expriment l'état de saturation du fragment concerné et sa capacité d'interaction avec d'autres fragments. La composition syntaxique consiste alors à  superposer partiellement ces fragments d'arbres pour saturer leurs polarités et obtenir un arbre unique complètement spécifié où toutes les polarités auront été saturées. 

On peut voir la composition syntaxique de façon totalement statique. L'ensemble des fragments d'arbres servant à construire un arbre syntaxique peut être vu comme une spécification d'une famille d'arbres qui constituent les modèles de cette spécification. C'est pourquoi nous l'appellerons une \emph{Description d'Arbre Polarisée (DAP)}. L'arbre syntaxique final représente alors un \emph{modèle} particulier de cette description.
La composition syntaxique apparaît ensuite comme  la réalisation d'une fonction d'interprétation  associant chaque nœud d'une DAP à un nœud d'un arbre syntaxique. On peut oublier le processus de composition pour ne conserver finalement que le triplet (DAP, arbre syntaxique, interprétation) que nous appellerons \emph{graphe d'interprétation}, dans la mesure où il peut se représenter sous forme d'un graphe. 

Seules les principales caractéristiques du formalisme des \ig~nécessaires à la compréhension de la suite de l'article sont données ici (voir \shortcite{Gui2010} pour une présentation complète).

Une DAP est un ensemble de nœuds représentant des syntagmes, structuré par des relations de domination et de précédence immédiates et sous-spécifiées. Les propriétés morpho-syntaxiques de chaque syntagme sont décrites par une structure de traits attachée à au nœud correspondant. Il existe deux types de traits:
\begin{itemize}
\item les traits \emph{polarisables} qui portent en plus de leur valeur une polarité qui peut être \emph{positif} (\positive), \emph{négatif} (\negative), \emph{virtuel} (\virtual) ou \emph{saturé} (\saturated)~; dans la suite deux traits de ce type seront utilisés~: \feat{cat} et \feat{funct}~;
\item les traits \emph{neutres} qui ne portent pas de polarités (le symbole $=$ est utilisé pour ces traits).
\end{itemize}
Dans la suite, les nœuds des modèles sont notés \mnode{N};  ceux des DAP, \inode{N}.

\begin{figure}[h]
\begin{center}
\includegraphics[scale=0.7]{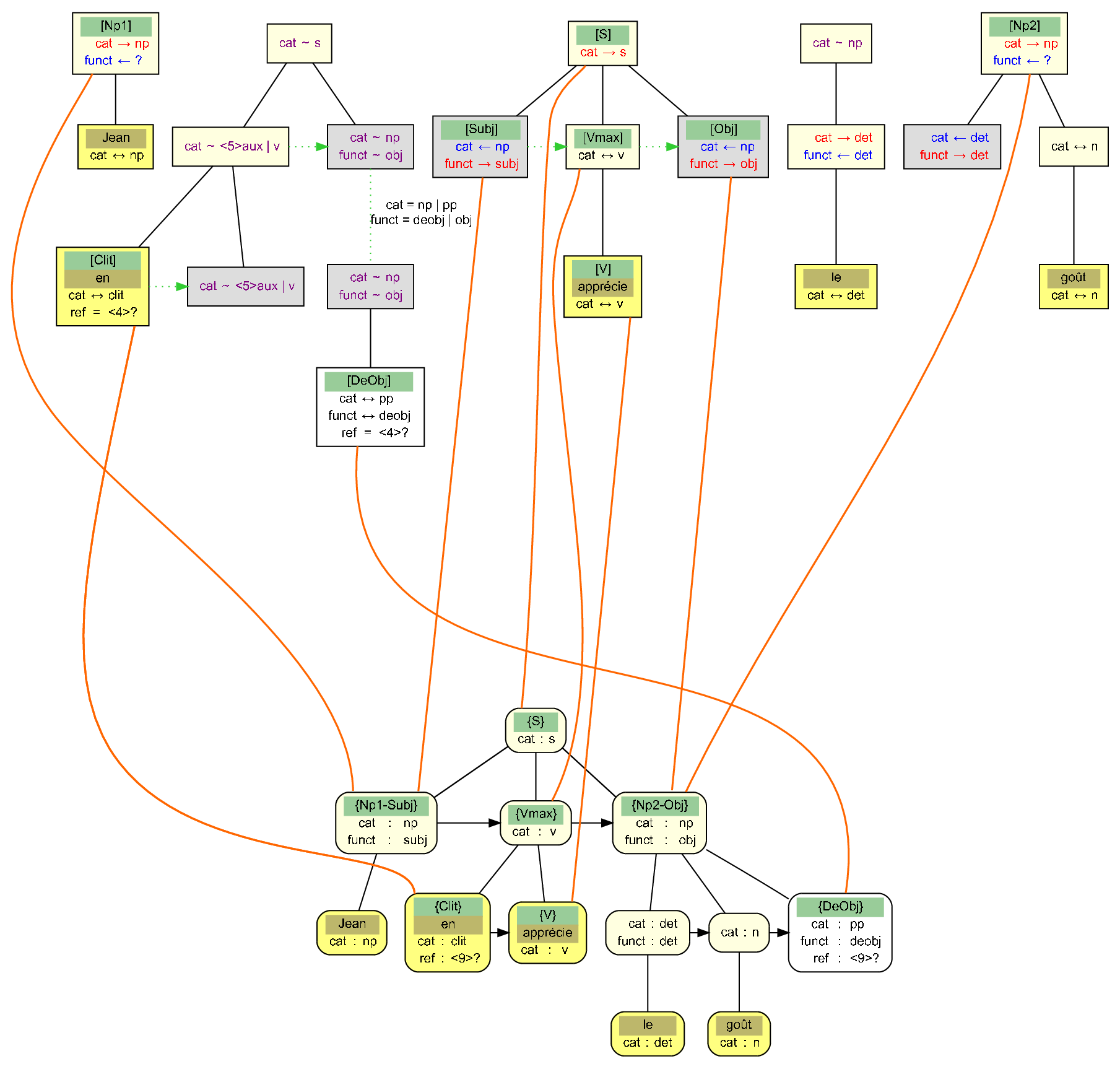}
\end{center}
\caption{\label{interpretation}Graphe d'interprétation de la phrase
  \french{Jean en apprécie le goût}}
\end{figure}

Une interprétation d'une DAP dans un arbre syntaxique est valide si elle préserve les relations de domination et de précédence. Par ailleurs, elle doit préserver les valeurs de traits ainsi que les relations de co-indexations entre celles-ci\footnote{A la différence de \shortcite{Gui2010}, nous considérons que les traits peuvent être co-indexés non seulement dans les DAP mais aussi dans les arbres syntaxiques.}.  Concernant la structure d'arbre ainsi que des traits étiquetant les nœuds, une interprétation garantit une minimalité du modèle en un sens défini dans \shortcite{Gui2010}.

Pour ce qui est des polarités, une interprétation valide doit respecter une des deux propriétés suivantes pour chaque ensemble de traits polarisés de la DAP de départ interprétés dans le même trait de l'arbre syntaxique d'arrivée~:
\begin{itemize}
\item {\bf cas non-linéaire~:} un seul trait est saturé et tous les autres sont virtuels;
\item {\bf cas linéaire~:} un trait est positif, un second négatif et tous les autres virtuels.
\end{itemize}
Une conséquence des conditions de saturation est que l'on peut définir,
pour un nœud \mnode{N} contenant un trait polarisable $f$,
\emph{l'antécédent principal} de \mnode{N} relativement à $f$
(noté $\rev{f}(\mnode{N})$) comme l'unique nœud de l'ensemble $\rev{\I} (\mnode{N})$ 
de la DAP porteur du trait $f$ saturé
(dans le cas non-linéaire) ou du trait $f$ positif (dans le cas linéaire).

Dans la suite, on appellera
\emph{nœud principal} un nœud d'une DAP qui porte un trait \feat{cat}
positif ou saturé (on notera donc {\tt cat}
$\rightarrow\mid\leftrightarrow$ {\tt ?} dans les motifs).

Une grammaire particulière d'interaction est définie par l'ensemble de ses DAP élémentaires (DAPE) utilisées pour composer des arbres syntaxiques.

Illustrons ces notions par l'exemple de l'analyse syntaxique de la phrase \french{Jean en apprécie le goût} avec la grammaire d'interaction du français \grf~\cite{perrier07}.  Dans une première phase, il s'agit de sélectionner les DAPE de \grf\ qui vont servir à analyser la phrase.  Elles sont réunies en une unique DAP $\D$ représentant le point de départ de l'analyse. Cette DAP est présentée par la figure~\ref{interpretation} dans sa partie haute. L'arbre syntaxique $\T$ résultant de l'analyse est présenté dans la partie basse de la même figure. La fonction d'interprétation de $\D$ dans $\T$ est représentée sur la figure par des arcs orange allant des nœuds de $\D$ vers ceux de $\T$\footnote{Elle est en fait partiellement représentée pour alléger la figure, la fonction d'interprétation étant en fait totale mais il n'est pas difficile de construire les arcs manquants.}. L'ensemble des deux structures et de la fonction d'interprétation de la figure~\ref{interpretation} constitue le graphe d'interprétation.

\vspace{-4mm}
\section{Principes de construction de la grammaire du français \grf}
\label{sec-prin}
\vspace{-2mm}

La grammaire \grf\ a été construite en suivant un certain nombre de règles qui sont l'expression dans le formalisme des \ig\ de principes  linguistiques qui ne sont pas spécifiques au français mais qui valent aussi pour d'autres langues plus ou moins proches.  Voici l'essentiel de ces règles~:
\begin{enumerate}
\item La grammaire est strictement \emph{lexicalisée}, ce qui signifie que chaque DAPE est associée à un mot-forme unique du français par le biais d'une feuille spéciale de la description, appelée son \emph{ancre}. Sur les figures, les ancres sont représentées en jaune foncé. L'ensemble des ascendants de l'ancre s'appelle l'\emph{épine dorsale} de la DAPE. 

\item Certains nœuds ont une forme phonologique vide. Ce sont toujours des feuilles qui représentent la trace d'un argument qui n'est pas dans sa position canonique. Cela peut correspondre à un argument extrait, un sujet inversé ou encore un clitique comme \french{en} dans notre exemple. Sur les figures, les nœuds vides sont représentés en blanc et les nœuds non vides en jaune~; dans les DAP, un nœud gris ne porte pas de contrainte sur la forme phonologique, il peut être vide ou non vide. Par exemple, sur la figure~\ref{interpretation}, la trace du complément de l'objet du verbe modifié par le clitique \french{en} est représentée par le nœud vide \inode{DeObj}.

\item Tous les nœuds de la grammaire portent un trait \feat{cat}. Pour chaque DAPE, tous les nœuds principaux non vides sont sur l'épine dorsale. Ces nœuds forment un chemin non vide commençant à un nœud que l'on appelle la \emph{projection maximum} de l'ancre et terminant à l'ancre elle-même. L'ancre est la \emph{tête} de tous ces nœuds et de façon duale, ceux-ci représentent ses diverses \emph{projections}. Pour une projection différente de l'ancre, on définit son \emph{fils principal} comme son fils qui est aussi une projection de la tête.

Sur la figure~\ref{interpretation}, dans la DAPE de \french{apprécie}, l'ancre \inode{V} a comme projections, outre elle-même, les nœuds \inode{Vmax} et \inode{S}. Dans la DAPE de \french{en}, l'ancre \inode{Clit} n'a qu'elle-même comme projection.

Telles qu'elles viennent d'être définies, les notions de tête et de projection sont relatives à une DAPE mais on peut les transposer à un arbre syntaxique modèle d'un ensemble de DAPE à l'aide d'une interprétation $\I$.  Pour tout nœud non vide \mnode{N}, $\rev{cat}(\mnode{N})$ est un nœud non vide d'une DAPE $D_i$ dont la tête est l'ancre \inode{A_i} de $D_i$. On dit alors que la tête de \mnode{N} est $\I(\inode{A_i})$ et que \mnode{N} est une projection de $\I(\inode{A_i})$.

Par exemple, dans l'arbre syntaxique $\T$ de la figure~\ref{interpretation}, le nœud \mnode{V} est la tête de \mnode{Vmax} et \mnode{S}.

\item Si un nœud d'un arbre syntaxique modèle d'une DAP est porteur d'un trait \feat{funct} avec une valeur $X$, cela signifie d'un point de vue linguistique que le syntagme correspondant remplit la fonction syntaxique $X$ par rapport à un syntagme représenté par un de ses nœuds frères dans l'arbre.

Par exemple dans l'arbre $\T$ de la figure~\ref{interpretation}, les nœuds \mnode{Subj} et \mnode{Obj} remplissent les fonctions respectives sujet et objet par rapport au noyau verbal représenté par leur frère \mnode{Vmax}.

Lorsqu'un nœud d'un arbre syntaxique pourvu d'un trait \feat{funct} de valeur $X$ a plusieurs frères, la lecture du modèle ne permet pas de déterminer lequel est celui par rapport auquel il remplit la fonction $X$. Pour cela, il faut revenir à la DAP correspondante via l'interprétation.  Nous devons distinguer trois cas. Considérons une DAP $\D$ composée de $n$ DAPE $\D_1, \ldots, \D_n$ qui est interprétée dans un modèle $\T$ via une interprétation $\I$. Considérons dans $\T$ un nœud \mnode{N} porteur d'un trait \feat{funct} de valeur $X$, le père de \mnode{N} étant noté \mnode{P}.

\begin{enumerate}
\item{\bfseries Interaction linéaire prédicat-argument.}  Le trait \feat{funct} est l'image d'un trait positif  issu d'une DAPE $\D_i$ et d'un trait négatif issue d'une autre DAPE  $\D_j$. Dans $\D_i$, la grammaire assure que le nœud \inode{N_i} porteur du trait \feat{funct} positif a toujours un unique frère \inode{M_i} qui est un nœud principal. Dans l'arbre $\T$, on peut alors dire que \mnode{N} remplit la fonction $X$ par rapport à l'image \mnode{M} de ce frère. On parle alors d'\emph{interaction linéaire} entre les DAPE  $\D_i$  et $\D_j$. Cette interaction est la réalisation d'une relation prédicat-argument.

Par exemple, il y a une interaction linéaire entre les DAPE associées à \french{goût} et à \french{apprécie} qui a pour résultat de réaliser la fonction objet du nœud \node{[Obj]} par rapport au nœud \node{[Vmax]}.

\item{\bfseries Interaction non-linéaire modifié-modificateur.} Le trait \feat{funct} est l'image d'un trait saturé issu d'une DAPE $\D_i$ et l'antécédent du nœud \mnode{N} dans $\D_i$ est un nœud \inode{N_i} qui a comme père un nœud \inode{P_i} pourvu d'un trait virtuel \feat{cat}. Il existe alors une DAPE unique $\D_j$ qui contient le nœud principal $\inode{P_j} = \rev{cat}(\mnode{P})$. Le fils principal \inode{M_j} de \inode{P_j} a pour image le frère \mnode{M} de \mnode{N}. Dans l'arbre $\T$, on peut alors dire que \mnode{N} remplit la fonction $X$ par rapport à \mnode{M}. On parle alors d'\emph{interaction non-linéaire} entre les DAPE $\D_i$ et $\D_j$. Cette interaction est la réalisation d'une relation de modification ou d'adjonction.

Par exemple, il y a une interaction non-linéaire entre les DAPE associées à \french{goût} et \french{en} qui a pour résultat de réaliser la fonction complément de nom du nœud \mnode{DeObj} par rapport au nœud \mnode{Np2-Obj}.

\item {\bfseries Relation prédicat-argument non réalisée.}  Le trait \feat{funct} est l'image d'un trait saturé issu d'une DAPE $\D_i$ et l'antécédent du nœud \mnode{N} dans $\D_i$ est un nœud vide \inode{N_i} qui a comme père un nœud principal \inode{P_i}. \inode{N_i}  a comme frère le fils principal \inode{M_i} de \inode{P_i}. Dans l'arbre $\T$, \mnode{N} remplit alors la fonction $X$ par rapport à l'image $\mnode{M}=\I(\inode{M_i})$ de ce frère.

Dans la figure~\ref{interpretation}, nous n'avons pas d'illustration de ce troisième cas que l'on rencontre en particulier pour représenter des relations prédicat-argument non réalisées phonologiquement, comme les relations verbe-sujet pour les infinitifs.
\end{enumerate}

\item \new{Si un nœud d'une DAPE porte un trait \feat{ref}, cela signifie que le syntagme correspondant est associé à une référence sémantique (la valeur du trait peut préciser la qualité de cette référence~: animée, inanimée mais concrète ou encore abstraite).  Si dans une même DAPE, deux nœuds ont des traits \feat{ref} co-indexés, cela signifie qu'ils renvoient à la même entité sémantique de référence. Par exemple, dans la DAPE associée à \french{en}, les nœuds \node{[Clit]} et \node{[DeObj]} ont des traits \feat{ref} co-indexés. Cela veut dire qu'ils représentent la même entité sémantique. De même, c'est avec la co-indexation de traits \feat{ref} que l'on modélise la différence de contrôle entre \french{permet} et \french{promet} (ce mécanisme s'apparente aux équations de contrôle de LFG).}

Cette co-indexation entre traits \feat{ref} de plusieurs nœuds se propage dans un modèle via la fonction d'interprétation et elle permet de réaliser des interactions indirectes entre syntagmes.
\end{enumerate}

\vspace{-4mm}
\section{Des motifs de graphe pour calculer les dépendances}
\label{sec-motifs} 
\vspace{-2mm}

Comme nous l'avons vu plus haut, pour calculer une structure en
dépendances, il est parfois nécessaire de considérer des informations
qui ne sont pas dans l'arbre syntaxique mais plutôt dans l'historique
de sa dérivation. En grammaire d'interaction, l'historique d'une
dérivation est décrit par ce que nous avons appelé le graphe
d'interprétation et qui représente le triplet (DAP, arbre syntaxique,
interprétation). Le calcul des dépendances à partir du graphe
d'interprétation peut alors s'exprimer à l'aide de motifs de graphe.

\vspace{-4mm}
\paragraph{Les motifs de graphe}
Un motif de graphe décrit un ensemble de contraintes à satisfaire par
le graphe d'interprétation pour qu'une dépendance soit produite.
Formellement, un motif de graphe est constitué d'un ensemble de motifs
de nœud et de relations entre ces motifs. Identifier un motif dans une
structure revient à construire une fonction d'appariement (on note
l'image de \node{N} par la fonction d'appariement $\app{N}$) qui
associe à chaque nœud du motif un nœud du graphe d'interprétation
compatible avec les contraintes exprimées par le motif. Il est
important de noter que les motifs font apparaître des nœuds de la DAP
(rectangles) et des nœuds du modèle (coins arrondis).

\begin{figure}[h]
  \centering
  \begin{tabular}{|M{2.9cm}|M{6.5cm}|M{6.5cm}|}
    \cline{2-3}
    \multicolumn{1}{c|}{}   
    & {\bf linéaire}\\ $\rev{funct}(\mnode{N})$ a un trait \feat{funct} positif 
    & {\bf non-linéaire}\\ $\rev{funct}(\mnode{N})$ a un trait \feat{funct} saturé
    \tabularnewline\hline
    {\bf canonique} \\ \mnode{N} est vide& 
    \includegraphics[scale=.4]{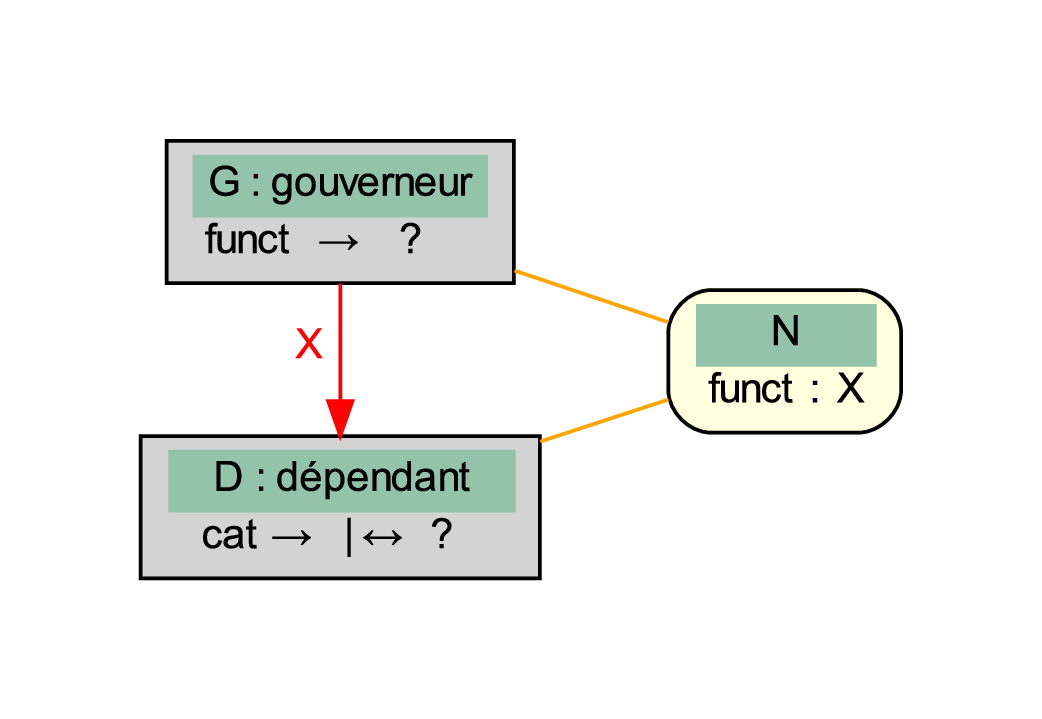} &  
    \includegraphics[scale=.4]{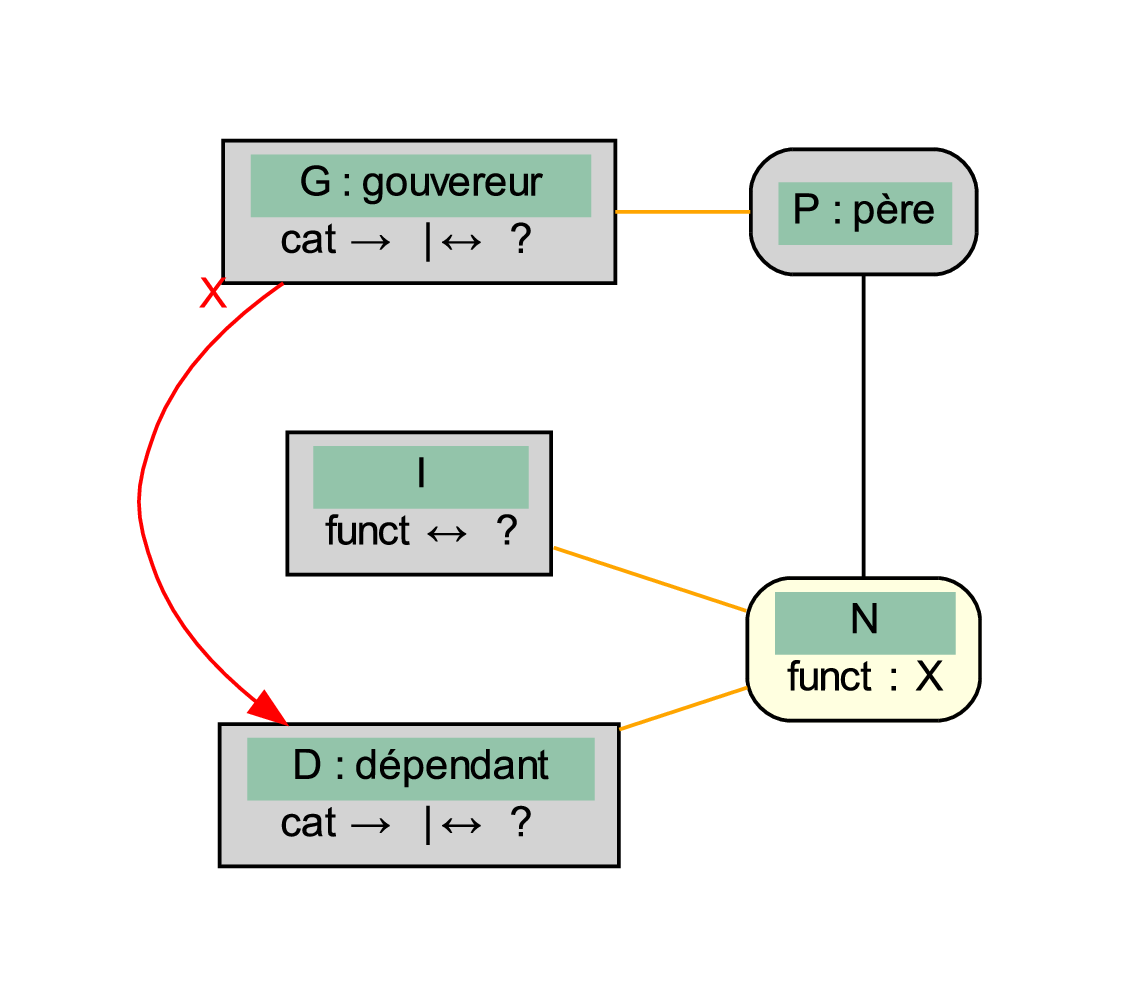}
    \tabularnewline\hline
    {\bf non-canonique}\\ \mnode{N} est non-vide &
    \includegraphics[scale=.4]{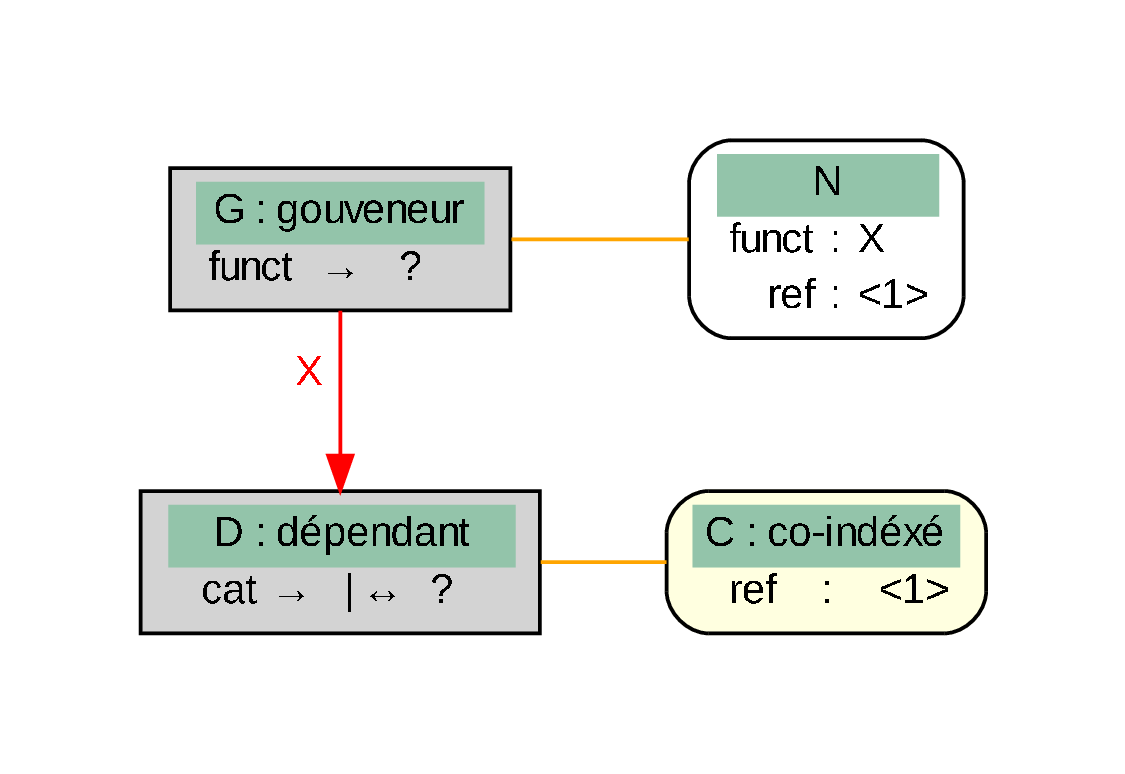} &
    \includegraphics[scale=.4]{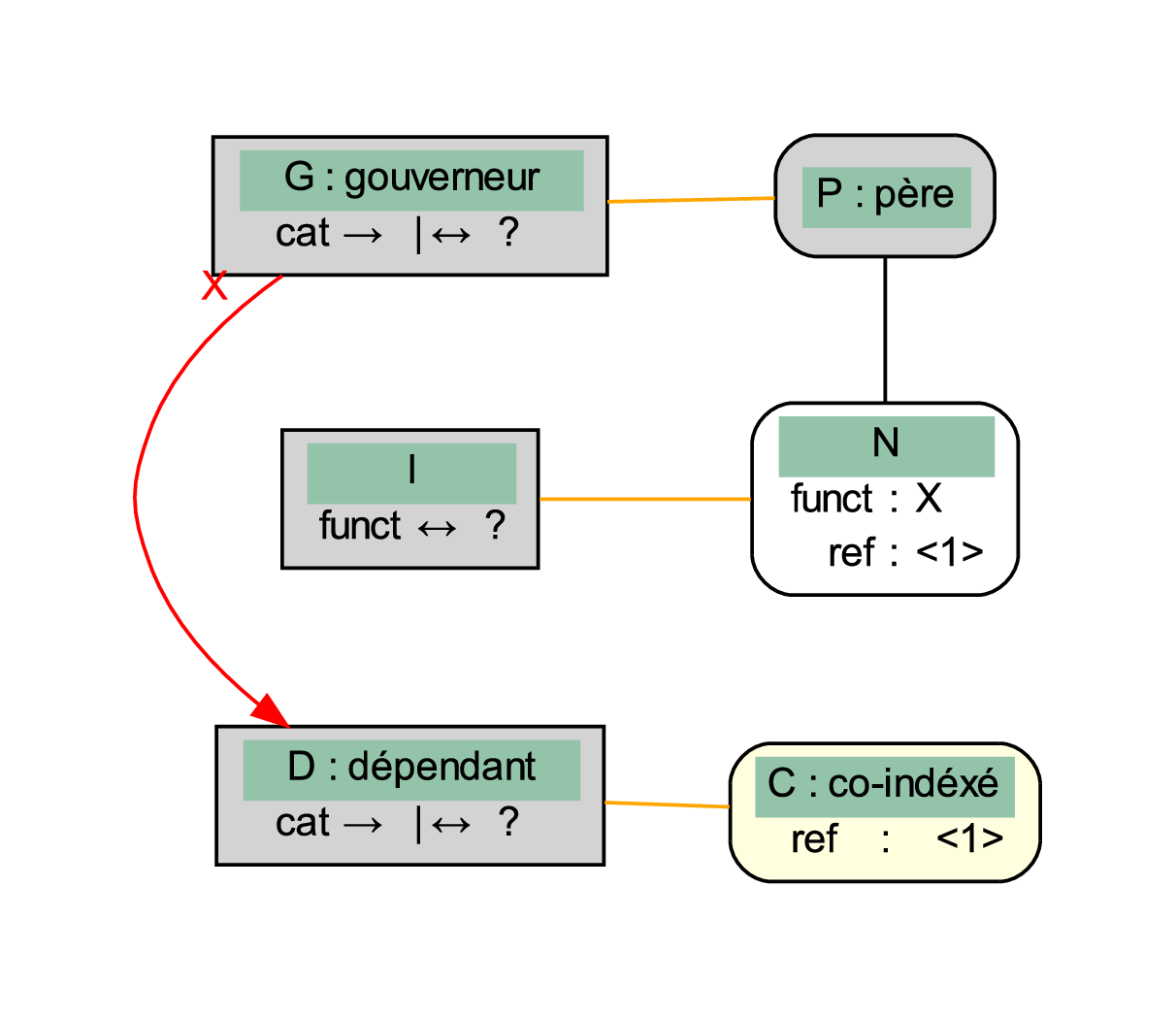} 
    \tabularnewline\hline
  \end{tabular}
\caption{\label{motifs} Motifs pour le calcul de dépendances}
\end{figure}

La figure~\ref{motifs} décrit les motifs que l'on utilise pour la grammaire \grf~; les contraintes portent 
sur les structures de trait, notamment sur les traits \feat{cat} et \feat{funct} et les polarités associées.
Ces contraintes portent également sur le fait qu'un nœud est vide (fond blanc) ou non vide (fond jaune) dans le modèle.

Pour les relations, deux types de contraintes sont utilisées. D'une part, on peut contraindre 
$\overline{\node{N}}$ à être l'interprétation de $\overline{\node{M}}$
(les motifs de nœuds \node{M} et \node{N} sont alors reliés par une arête orange dans le motif).
D'autre part, on peut contraindre $\overline{\node{N}}$ à être un sous-constituant
  immédiat de $\overline{\node{M}}$ (\node{M}  est au-dessus de \node{N} et ils sont
  reliés par un trait noir).

Chaque motif décrit un ensemble de contraintes à vérifier pour qu'une dépendance soit ajoutée.
 La flèche rouge ne fait pas partie du
motif, elle indique simplement qu'une dépendance doit être ajoutée
quand le motif est repéré dans le graphe d'interprétation~; la
dépendance créée relie alors les mots-formes portés par les ancres des
descriptions correspondant aux nœuds \node{G} et \node{D}.

\begin{figure}[h]
  \centering
  \includegraphics[scale=.8]{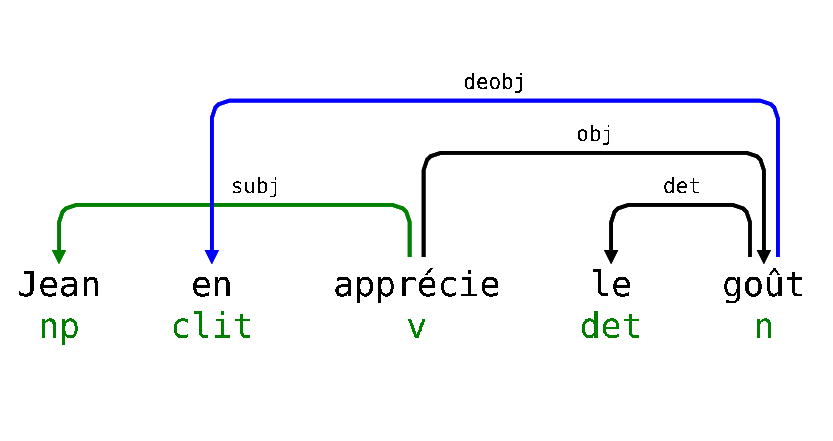} 
\caption{Structure en dépendance pour la phrase \french{Jean en apprécie le goût}}
  \label{fig:dep}
\end{figure}

Par exemple, on peut appliquer le motif représentant le cas linéaire
et canonique, en haut et à gauche dans la figure \ref{motifs}, au
graphe d'interprétation de la figure~\ref{interpretation} par
l'appariement~: $\app{N} = \mnode{Np1-Subj}$, $\app{G} = \inode{Subj}$
et $\app{D} = \inode{Np1}$. On vérifie facilement que toutes les
contraintes imposées par le motif sont vérifiées~; on peut donc
ajouter une relation de dépendance portant l'étiquette \feat{subj}
(valeur du trait \feat{funct} dans le nœud \app{N}) entre
\french{apprécie} (mot-forme de l'ancre de la DAPE qui contient le
nœud $\app{G}$) et \french{Jean} (mot-forme de l'ancre de la DAPE qui
contient le nœud $\app{D}$). Cela correspond à la dépendance verte
dans la figure~\ref{fig:dep}.

\vspace{-4mm}
\paragraph{Motifs de graphe pour les dépendances complètes}
Présentons maintenant les quatre motifs qui s'appuient
sur les principes de la grammaire pour calculer les dépendances
complètes d'une phrase. La grammaire modélise chaque dépendance par
l'utilisation d'un trait {\tt funct}~; il s'agit donc d'interpréter
les principes décrits dans le point~4 de la section~\ref{sec-prin} de
telle façon que:
\vspace{-4mm}
\begin{center}
\emph{si \mnode{N} remplit la fonction syntaxique $X$ par rapport à un
  frère \mnode{M}\\ alors une dépendance existe entre la tête de \mnode{M}
  et la tête de \mnode{N}.}
 \end{center}
\vspace{-3mm}

 Les quatre règles de la figure~\ref{motifs} contiennent toutes un
 motif de n\oe ud \node{N} avec le trait \feat{funct} de valeur $X$.
 Elles correspondent à la
 combinaison de deux alternatives~: la linéarité et le fait que le
 dépendant soit en position canonique ou pas. Pour chaque nœud
 \mnode{N} du modèle portant un trait funct de valeur $X$, on fixe
 $\app{N}=\mnode{N}$ et on distingue~:

\begin{description}
\item[Le cas linéaire~:] ce cas correspond aux deux règles à gauche
  dans la figure~\ref{motifs} et il est caractérisé par le fait que
  $\rev{funct}(\mnode{N})$ a un trait \feat{funct} positif. Cela
  correspond au point 4(a) de la section~\ref{sec-prin} et donc, le
  nœud par rapport auquel \mnode{N} remplit la fonction syntaxique
  $X$ est dans la même DAPE que $\rev{funct}(\mnode{N})$ et donc $\app{G} =
  \rev{funct}(\mnode{N})$.

  \item[Le cas non-linéaire~:] ce cas (les deux règles de droite)
  s'applique quand $\rev{funct}(\mnode{N})$ a un trait \feat{funct}
  saturé (4(b) et 4(c) de la section~\ref{sec-prin}). Le nœud
  \mnode{M} par rapport auquel \mnode{N} remplit la fonction
  syntaxique $X$ est le fils principal du nœud \inode{P_j} dans le
  cas 4(b) et du nœud \inode{P_i} dans le cas 4(c). Dans les deux
  cas, ce nœud \mnode{M} est donc dans la même DAPE que la tête du
  père \mnode{P} du nœud \mnode{N}.
  
\item[Le cas canonique~:] le dépendant de la relation de dépendance
  est la tête du nœud \mnode{N} quand elle existe (c'est-à-dire
  quand \mnode{N} est non-vide) et donc par définition cette tête est
  dans la même DAPE que $\rev{cat}(\mnode{N})$ c'est le cas pour les
  deux motifs de graphe en haut de la figure~\ref{motifs} qui
  correspondent au cas où le dépendant est en position canonique.

\item[Le cas non-canonique~:] si le nœud \mnode{N} est vide, on
  utilise le principe du point 5 de la partie~\ref{sec-prin}~; ce
  principe assure qu'un nœud \mnode{C} non-vide dont le
  trait \feat{ref} est co-indexé avec celui du nœud \mnode{N}
  renvoie à la même entité sémantique, c'est donc ce nœud qui a
  pour tête le dépendant réel~;  les deux motifs de graphe du bas de la
  figure~\ref{motifs} s'appliquent alors avec $\app{C} = \mnode{C}$ et
  $\app{D} =\rev{cat}(\mnode{C})$. L'existence et l'unicité d'un tel
  nœud \mnode{C} est assuré par la grammaire.
\end{description}

Par exemple, considérons le trait $\feat{funct}:\feat{deobj}$ du nœud
\mnode{DeObj} de la figure \ref{interpretation}.

\setlength{\extrarowheight}{2pt}
\tabletail{\hline}
\begin{supertabular}[c]{|m{11cm}|m{6cm}|}
\hline
On considère  le trait $\feat{funct}:\feat{deobj}$ du nœud \mnode{DeObj}  & 
$\app{N}=\mnode{DeObj}$  \tabularnewline \hline
$\rev{funct}(\mnode{Deobj}) = \inode{DeObj}$ qui a un trait \feat{funct}
saturé donc cas {\bf non-linéaire} & 
$\app{I} =\inode{DeObj}$  \tabularnewline \hline
le père de \mnode{DeObj} est \mnode{Np2-Obj}
 & $\app{P} =\mnode{Np2-Obj}$  \tabularnewline \hline
$\rev{cat}(\mnode{Np2-Obj}) = \inode{Np2}$
& $\app{G} =\inode{Np2}$  \tabularnewline \hline
\mnode{DeObj} est vide (cas {\bf non-canonique}), on considère l'unique 
nœud non-vide avec le même index $\langle9\rangle$, il s'agit de \mnode{Clit}
& $\app{C} =\mnode{Clit}$  \tabularnewline \hline
$\rev{cat}(\mnode{Clit}) = \inode{Clit} $
 & $\overline{\node{D}} =\inode{Clit}$  \tabularnewline \hline
\end{supertabular}

Le motif de graphe pour le cas {\bf non-linéaire non-canonique}
s'applique donc, ce qui donne la dépendance (dessinée en bleu sur la
figure~\ref{fig:dep}) \feat{deobj} entre \french{goût} (le mot-forme
de l'ancre de $\app{G} =\inode{Np2})$ et \french{en} (le mot-forme de
l'ancre de $\app{D} = \inode{Clit})$. Sur la figure~\ref{fig:dep}, les
trois autres dépendances sont des applications du cas linéaire canonique.\footnote{D'autres exemples de structures de dépendances obtenues par la méthode décrite ci-dessus peuvent être consultés à l'adresse~\url{http://leopar.loria.fr/exemples_dep/}.}.

\vspace{-4mm}
\section{Conclusion}
\vspace{-2mm}

Nous avons présenté une méthode pour calculer les dépendances
syntaxiques d'un énoncé à partir du processus d'analyse en
constituants à l'aide des grammaire d'interaction. Cette méthode à
base de motifs de graphes permet de retranscrire tout l'information de
l'analyse en constituants nécessaire à la construction de la
sémantique. Il nous reste maintenant à valider notre méthode sur des
corpus à grande échelle, par exemple dans le cadre d'une campagne
d'évaluation comme PASSAGE.

D'autre part, notre méthode de sélection de motifs de graphe peut être
généralisée pour l'analyse sémantique. Il ne s'agit plus de
détecter des motifs de graphes mais d'appliquer des
transformations directement sur les graphes dans le cadre de
la réécriture de graphes~\cite{taln_beta}.

\bibliographystyle{taln2002}
\bibliography{ig}

\end{document}